\documentclass{article}

\usepackage{microtype}
\usepackage{graphicx}
\usepackage{subcaption}
\usepackage{booktabs} 

\usepackage{hyperref}

\usepackage[preprint]{icml2026}

\usepackage{amsmath}
\usepackage{amssymb}
\usepackage{mathtools}
\usepackage{amsthm}

\icmltitlerunning{Making Time Editable in Video Diffusion Transformers}

\begin{document}

\twocolumn[
\icmltitle{Making Time Editable in Video Diffusion Transformers}

\begin{icmlauthorlist}
\icmlauthor{Konstantin Kuklev}{aff1}
\icmlauthor{Viacheslav Vasilev }{aff1}
\icmlauthor{Alexander Kunitsyn}{aff1}
\icmlauthor{Andrei Ivaniuta}{aff1}
\icmlauthor{Denis Dimitrov }{aff1}
\end{icmlauthorlist}

\icmlaffiliation{aff1}{Kandinsky Lab}

\icmlcorrespondingauthor{Andrei Ivaniuta}
{andrei.ivaniuta@kandinskylab.ai}
\icmlcorrespondingauthor{Konstantin Kuklev}{konstantin.kuklev@kandinskylab.ai}


\icmlkeywords{diffusion transformers, video generation, temporal conditioning}

\vskip 0.3in
]

\printAffiliationsAndNotice{} 

\begin{abstract}
Modern Diffusion Transformers for video generation provide limited control over the progression of time and the editing of temporal dynamics. We propose a temporal-control methodology that extends a pretrained DiT with explicit time editing, allowing control over motion speed and temporal structure without redesigning the backbone. Its core implementation augments the pretrained model with a lightweight temporal module, preserving the original generative prior while expanding its controllable dynamic range.
\end{abstract}

\section{Introduction}

Modern Diffusion Transformer (DiT) architectures have recently achieved strong performance in video generation, producing high-fidelity frames and realistic motion across a wide range of scenarios~\cite{lu2023vdt,latte,wan}. However, despite this progress, temporal controllability can remain limited in practice.

Maintaining consistent and controllable temporal dynamics is challenging. Temporal dependencies can be entangled with content and motion representations and are only indirectly influenced through conditioning or training dynamics, rather than being modeled as explicit, independently controllable variables~\cite{yu2024cmd,kim2025ticft}.

We argue that this limitation arises because time is overloaded. In standard DiT-based video generation, diffusion time may act as a proxy for both denoising progress and motion evolution. As a result, the model does not explicitly learn an independently controllable representation of temporal evolution. Instead, temporal behaviour is absorbed into a single entangled variable, creating ambiguity between \emph{how far denoising has progressed} and \emph{how motion should unfold in time}.

\begin{figure}[t]
    \centering
    \includegraphics[height=0.33\textheight]{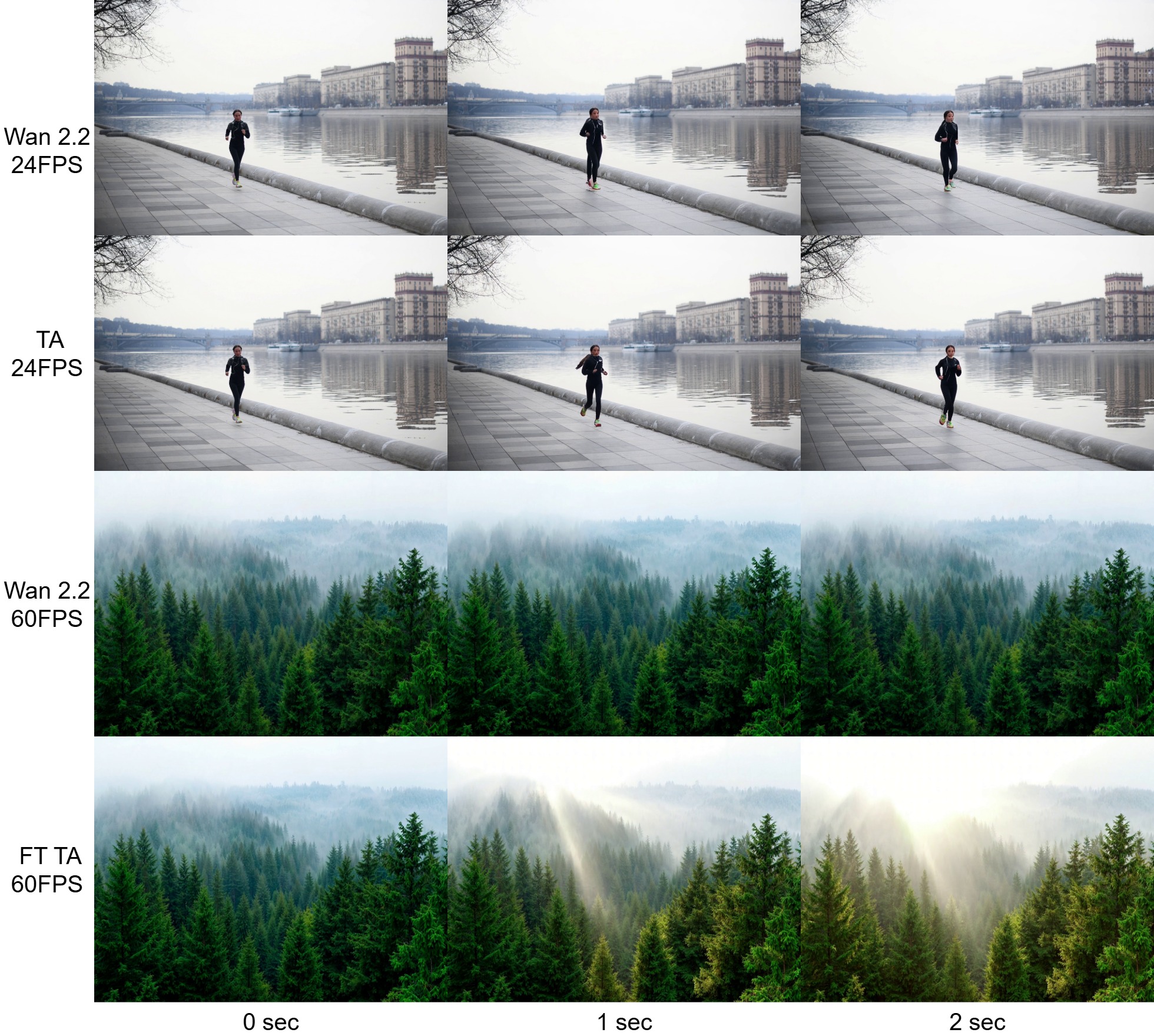}
    \caption{Qualitative comparison of temporal control across two prompts. 
Rows 1–2 show the prompt ``A girl runs along the embankment'' generated at 24 FPS with Wan2.2 (row 1) and Wan2.2 + TA (row 2).
The model with TA produces smoother intermediate motion with fewer artifacts in the arms and legs, indicating improved temporal consistency even at standard playback speeds.
Rows 3–4 show the natural-process prompt ``Sunlight begins to illuminate the forest, piercing through the fog'' using FTTA. 
The generated sequence demonstrates gradual dissipation of the fog over time, indicating that explicit temporal control affects the physical evolution of the scene rather than only its playback rate.}
    \label{fig:ta_compare}
\end{figure}

This ambiguity becomes particularly visible when the temporal grid is changed relative to training. A pretrained DiT may still produce plausible frames, but struggles to maintain consistent temporal dynamics under changes in target FPS (frames per second) or motion pacing, as illustrated in Figure~\ref{fig:ta_compare} (rows 3--4).

This reflects a deeper limitation: the model does not reliably capture how temporal discretization relates to the physical evolution of motion. As a result, intermediate dynamics may become degraded (row~1), and slow processes such as gradual changes in the scene may be underrepresented or fail to evolve (row~3). The model preserves appearance priors well, but lacks a consistent understanding of how motion should unfold over time.

\newpage

Existing controllable video generation methods address parts of this problem. They can control motion, trajectories, object placement, or camera behaviour. However, they do not directly make time itself an explicitly editable variable. They shape \emph{what} moves, and sometimes \emph{where} it moves, but not cleanly \emph{how temporal dynamics are factorized and manipulated}. See more detailed review in Appendix~\ref{sec:related_works}.

\textbf{Contribution.} We propose a method that moves DiT-based video models closer to temporal factorization. Our approach extends a pretrained DiT backbone with an explicit temporal control module that enables more direct control over time and scene dynamics while preserving the strengths of the original generator. Thus, we retain the visual and structural priors of the pretrained model, but make temporal progression itself a controllable variable rather than an implicit side effect of denoising. This shift makes temporal editing more reliable and structured, and can also improve temporal consistency and motion coherence under standard generation settings, as illustrated in Figure~\ref{fig:ta_compare}, row~2 and 4.
We further observe that the same temporal-control behaviour transfers across multiple DiT backbones with different parameter scales and output resolutions, suggesting that explicit temporal factorization is not specific to a single architecture.


\section{Method}
\begin{figure}
    \centering
    \includegraphics[width=0.99\linewidth]{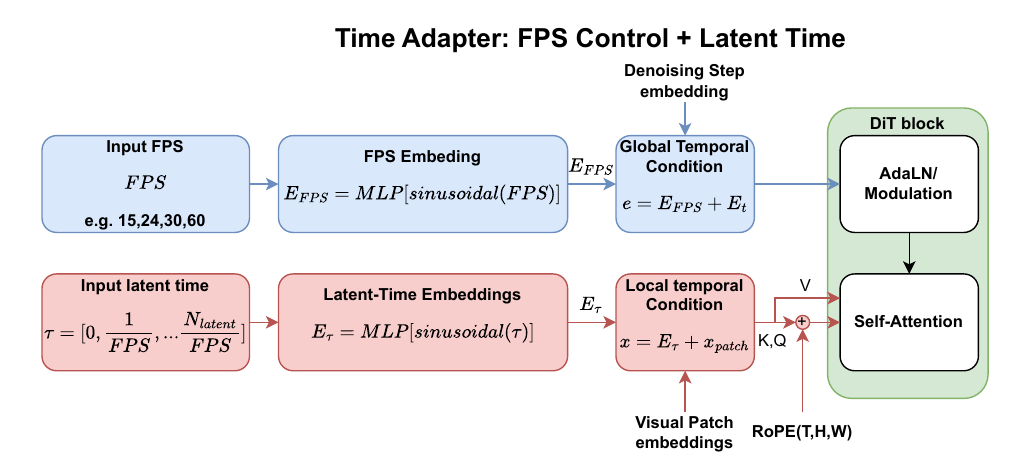}
    \caption{Time Adapter architecture for explicit temporal control in video Diffusion Transformers. FPS conditioning modulates global motion rate, while latent-time embeddings align local temporal progression. TA trains only temporal branches, whereas FTTA additionally fine-tunes the DiT backbone.}
    \label{fig:ta_structure}
\end{figure}

\subsection{Overview}

Our method extends a pretrained DiT backbone with an explicit temporal control module that separates global motion pacing from local temporal alignment, as illustrated in Figure~\ref{fig:ta_structure}. 
We introduce two complementary temporal signals: a global FPS embedding controlling motion rate and a latent-time embedding aligning frame-wise temporal progression. 
During training, the model is exposed to frame sequences with varying temporal densities through FPS conditioning, encouraging more consistent relationships between frames and improving intermediate motion quality.

Temporal conditioning is factorized at two levels of the architecture:
\[
e = E_t(t) + E_{\mathrm{fps}}(\mathrm{fps}),
\qquad
\]
\[
h_0 = \mathrm{PatchEmbed}(x_t) + E_{\tau}(\tau),
\]

where $E_{\mathrm{fps}}$ and $E_{\tau}$ serve complementary roles. 
The FPS embedding modulates the global motion-rate regime through the denoising conditioning pathway and determines the expected pacing of scene evolution. 
In contrast, latent-time embeddings provide frame-level temporal correspondence and local temporal ordering through token conditioning. 

Together they separate global motion pacing from local temporal alignment, allowing temporal progression to be represented more explicitly rather than relying on diffusion time alone as an implicit proxy for motion evolution.
Unlike RoPE~\cite{su2023roformerenhancedtransformerrotary}, which provides relative spatio-temporal token geometry inside attention, latent-time embeddings introduce an explicit temporal coordinate that remains editable during generation. The two mechanisms are therefore complementary rather than redundant.

The temporal modules are integrated through lightweight additive adapters initialized to preserve the pretrained backbone behaviour. TA trains only the temporal branches, whereas FTTA additionally fine-tunes the full DiT backbone. A detailed formulation and training objective are provided in 
Appendix~\ref{app:formulation}.
Training uses a curated dataset of approximately 40k videos spanning human actions and natural processes across multiple FPS regimes (24--60 FPS), providing coverage of both articulated motion and gradual scene evolution.

\begin{figure*}[t]
    \centering
    \includegraphics[width=0.85\textwidth]{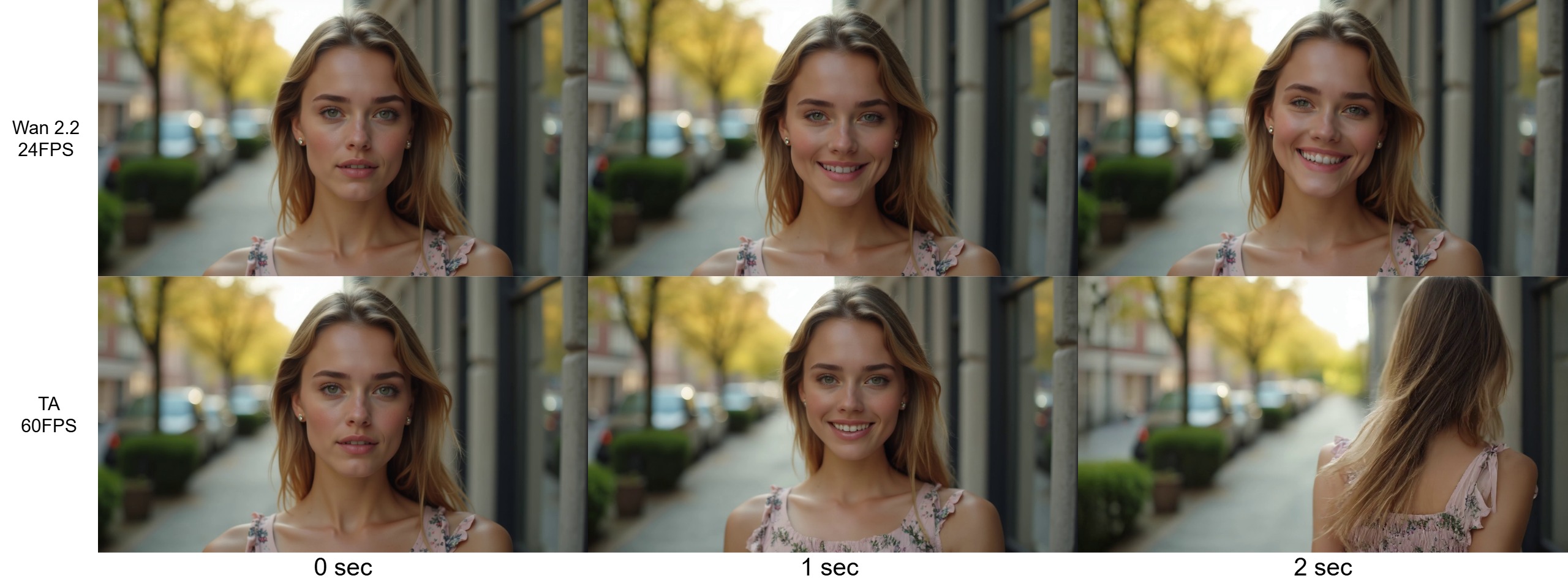}
    \caption{
    Frames at identical timestamps (0, 1, 2 s) from videos generated with different FPS using Wan2.2 and Wan2.2 + TA for the prompt ``Close-up of a girl in a dress; she smiles at the camera, then turns around and leaves.'' Increasing FPS from 24 to 60 yields approximately 2$\times$ faster motion, indicating precise and physically consistent control of temporal dynamics.}

    \label{fig:speed_diff_frames}
\end{figure*}
\section{Evaluation Protocol}

Our evaluation follows VMBench~\cite{vmbench2025} on two domains: Human Actions and Natural Processes. Metric definitions and benchmark details are provided in Appendix.~\ref{app:benchmarks}


\begin{table*}[t]
\centering
\scriptsize
\setlength{\tabcolsep}{3.5pt}
\caption{Benchmark results across domains, models, and target FPS. Lower is better for Flow Acceleration and LPIPS-Var; higher is better for PAS, MSS, and FDC.}
\label{tab:vmbench-results}
\resizebox{\textwidth}{!}{%
\begin{tabular}{lrrrrr rrrrr rrrrr rrrrr}
\toprule
Model
& \multicolumn{5}{c}{15 FPS}
& \multicolumn{5}{c}{24 FPS}
& \multicolumn{5}{c}{30 FPS}
& \multicolumn{5}{c}{60 FPS} \\
\cmidrule(lr){2-6}\cmidrule(lr){7-11}\cmidrule(lr){12-16}\cmidrule(lr){17-21}
& PAS & MSS & Accel. & LPIPS-V. & FDC
& PAS & MSS & Accel. & LPIPS-V. & FDC
& PAS & MSS & Accel. & LPIPS-V. & FDC
& PAS & MSS & Accel. & LPIPS-V. & FDC \\
\midrule
\multicolumn{21}{c}{\textbf{Human Actions}} \\
\midrule
Baseline Wan2.2
& \textbf{0.277} & \textbf{0.836} & 0.394 & 0.0319 & 0.831
& \textbf{0.293} & 0.751 & 0.474 & 0.0337 & 0.852
& \textbf{0.230} & \textbf{0.786} & 0.495 & 0.0309 & 0.859
& \textbf{0.292} & \textbf{0.815} & 0.408 & 0.0315 & 0.869 \\
TA Wan2.2
& 0.126 & 0.785 & 0.360 & 0.0376 & 0.656
& 0.207 & \textbf{0.762} & 0.170 & 0.0217 & \textbf{0.869}
& 0.116 & 0.752 & 0.223 & 0.0223 & \textbf{0.906}
& 0.230 & 0.726 & 0.226 & 0.0211 & \textbf{0.887} \\
FTTA Wan2.2
& 0.136 & 0.740 & \textbf{0.085} & \textbf{0.0178} & \textbf{0.851}
& 0.078 & 0.646 & \textbf{0.068} & \textbf{0.0174} & 0.796
& 0.067 & 0.721 & \textbf{0.076} & \textbf{0.0155} & 0.834
& 0.128 & 0.727 & \textbf{0.082} & \textbf{0.0180} & 0.847 \\
\midrule
\multicolumn{21}{c}{\textbf{Natural Processes}} \\
\midrule
Baseline Wan2.2
& -- & \textbf{0.993} & 0.049 & 0.0069 & \textbf{0.488}
& -- & 0.935 & 0.057 & 0.0059 & \textbf{0.635}
& -- & 0.987 & 0.050 & 0.0079 & 0.471
& -- & 0.995 & 0.057 & 0.0073 & 0.399 \\
TA Wan2.2
& -- & 0.983 & 0.199 & 0.0199 & 0.334
& -- & 0.968 & 0.021 & \textbf{0.0037} & 0.423
& -- & \textbf{0.998} & 0.023 & 0.0061 & 0.578
& -- & 0.986 & \textbf{0.018} & 0.0055 & 0.228 \\
FTTA Wan2.2
& -- & 0.992 & \textbf{0.032} & \textbf{0.0049} & 0.327
& -- & \textbf{0.998} & \textbf{0.020} & 0.0043 & 0.509
& -- & 0.995 & \textbf{0.015} & \textbf{0.0046} & \textbf{0.581}
& -- & \textbf{0.998} & 0.020 & \textbf{0.0041} & \textbf{0.624} \\
\bottomrule
\end{tabular}%
}
\end{table*}
\section{Experiments}

We conduct experiments on Wan2.2~\cite{wan}, which serves as the primary benchmark backbone due to its stable video generation across multiple FPS regimes. We compare two training regimes for temporal control: the proposed Time Adapter (TA), which preserves the frozen backbone and injects explicit FPS and latent-time conditioning, and Fine-Tuned Time Adapter (FTTA), which uses the same temporal conditioning together with full backbone fine-tuning. This setup allows us to compare explicit temporal conditioning against stronger global temporal regularization.

As an additional test of the proposed temporal-control methodology, we apply it to Kandinsky 5.0 Lite, a high-quality open-source video generation model~\cite{kandinsky5}, to assess whether the advantages of explicit time editing transfer beyond the Wan2.2 backbone.


\section{Results}

Evaluating temporal dynamics in video generation remains challenging, as standard metrics primarily capture motion magnitude or perceptual similarity rather than the quality of intermediate motion or the physical consistency of temporal evolution. Artifacts in articulated motion and failures in gradual processes are often not reliably detected. In some cases, stronger temporal smoothing can even increase smoothness scores (e.g., MSS) while perceptual quality deteriorates. We therefore complement quantitative evaluation with qualitative analysis.

Table~\ref{tab:vmbench-results} summarizes the benchmark results across the two evaluation domains and four target FPS settings. For Natural Processes, PAS is omitted because the setup contains no explicit foreground subject. Consistent with the limitations above, improvements in temporal dynamics are not always reflected by existing metrics (see Figure~\ref{fig:ta_compare}).

In particular, PAS should be interpreted as a motion-amplitude diagnostic rather than a direct measure of motion realism. As a result, articulated artifacts may increase PAS despite reducing perceived motion quality, partially explaining the discrepancy between quantitative metrics and human evaluation.

\paragraph{Human Actions.}
On Human Actions, both TA and FTTA reduce Flow Acceleration and LPIPS-Var, producing more stable short-term temporal transitions than the baseline. TA achieves the strongest directional coherence (FDC) at 24--60 FPS, suggesting that explicit temporal conditioning improves intermediate motion consistency when the target FPS approaches the training regime.

We hypothesize that exposure to denser temporal transitions improves intermediate motion interpolation and reduces perceptual artifacts, even at standard playback speeds. FTTA produces the smoothest dynamics overall but often generates motion that appears overly conservative in articulated human actions, whereas TA better preserves the pretrained motion prior while improving temporal consistency.

\paragraph{Natural Processes.}
Natural Processes exhibit a different trend. FTTA consistently achieves the strongest temporal regularization, obtaining the best MSS and the lowest Flow Acceleration and LPIPS-Var in most settings.

Qualitative results explain this behavior. In several Natural Processes examples, the baseline produces weak or nearly static dynamics, while FTTA makes scene evolution more responsive to temporal conditioning. This stronger temporal restructuring is beneficial for distributed physical processes such as fog, waves, clouds, and precipitation.
\begin{table}[t]
\centering
\caption{Side-by-side human evaluation against Wan2.2. Values denote win rates of TA or FTTA over the baseline; values above 0.5 favor the time-controlled model.}
\label{tab:sbs_criteria_benchmarks}
\small
\begin{tabular}{lcccc}
\toprule
& \multicolumn{2}{c}{Human Actions} & \multicolumn{2}{c}{Natural Processes} \\
\cmidrule(lr){2-3}\cmidrule(lr){4-5}
Criterion & TA & FTTA & TA & FTTA \\
\midrule
Naturalness & \textbf{0.58} & 0.45 & 0.43 & \textbf{0.50} \\
Speed consistency & \textbf{0.66} & \textbf{0.79} & \textbf{0.63} & \textbf{0.76} \\
Prompt fidelity & 0.45 & 0.37 & 0.43 & \textbf{0.69} \\
\bottomrule
\end{tabular}
\end{table}

\paragraph{Side-by-Side Evaluation.}
The SBS study in Table~\ref{tab:sbs_criteria_benchmarks} supports the same interpretation. TA is generally preferred for articulated human motion and motion naturalness, whereas FTTA achieves stronger perceived temporal control for distributed physical processes.

The SBS results also highlight an important distinction between explicit temporal control and FPS resampling. Unlike post-hoc playback modification, the proposed method changes the generated temporal evolution itself. As illustrated in Figure~\ref{fig:speed_diff_frames}, increasing the target FPS produces faster scene progression rather than simple frame interpolation.

Table~\ref{tab:sbs_k5} shows that the same qualitative trend is also observed on Kandinsky 5.0 Lite despite substantial differences in backbone scale and output resolution, providing preliminary evidence that explicit temporal factorization may transfer across different DiT backbones.

\label{app:K5SBS}
\begin{table}[t]
\centering
\caption{Side-by-side human evaluation against Kandinsky 5.0 Lite. Values denote win rates for TA or FTTA over Kandinsky 5.0 Lite; values above 0.5 favor the time-controlled model.}
\label{tab:sbs_k5}
\small
\begin{tabular}{lcccc}
\toprule
& \multicolumn{2}{c}{Human Actions} & \multicolumn{2}{c}{Natural Processes} \\
\cmidrule(lr){2-3}\cmidrule(lr){4-5}
Criterion & TA & FTTA & TA & FTTA \\
\midrule
Naturalness & \textbf{0.77} & 0.70 & 0.50 & \textbf{0.55} \\
Speed consistency & \textbf{0.66} & 0.65 & 0.58 & \textbf{0.68} \\
Prompt fidelity & \textbf{0.75} & 0.53 & \textbf{0.80} & 0.69 \\
\bottomrule
\end{tabular}
\end{table}
\section{Ablation Study}


\paragraph{Ablation Study.}
We study three aspects of the proposed temporal-control design. First, we compare the baseline DiT against TA to evaluate whether explicit temporal conditioning can be introduced while preserving the pretrained motion prior. Second, we compare TA against FTTA to separate lightweight temporal control from full-model temporal regularization. The results show that stronger fine-tuning is more beneficial for Natural Processes than for Human Actions.

Third, we ablate the two temporal signals used by the method: global FPS conditioning and local latent-time embeddings. The global FPS signal controls coarse motion pacing, while latent-time embeddings provide local temporal alignment. Table~\ref{tab:ablation_factorization} shows that FPS conditioning alone remains closest to the baseline in PAS, whereas latent-time conditioning contributes more strongly to local temporal consistency. The full model achieves the strongest reduction in Flow Acceleration, suggesting that stable temporal dynamics require both global motion pacing and local temporal alignment.

\begin{table}[t]
\centering
\caption{Ablation of temporal factorization into global FPS conditioning and local latent-time embeddings.}
\label{tab:ablation_factorization}
\small
\begin{tabular}{lcccc}
\toprule
Variant & PAS & FA & LPIPS-Var & Comment \\
\midrule
Baseline & \textbf{0.273} & 0.443 & 0.0320 & no control \\
FPS only & 0.248 & 0.415 & \textbf{0.0034} & global pacing \\
Lat time only & 0.175 & 0.363 & 0.0037 & local alignment \\
Full (ours) & 0.170 & \textbf{0.245} & 0.0257 & smooth control \\
\bottomrule
\end{tabular}
\end{table}

\section{Limitations}

Our approach exposes two fundamental limitations of current video diffusion models.

First, temporal factorization improves controllability within the training regime, but remains sensitive to underrepresented temporal scales. In particular, low-FPS regimes reveal that explicit time representations cannot compensate for missing temporal priors in the data. This suggests that controllable time requires not only better parameterization, but also broader coverage of temporal dynamics during training.

Second, temporal editability is inherently bounded by the fixed generation horizon of diffusion models. While the proposed method makes temporal dynamics more controllable, it does not increase the number of generated frames. As a result, strong temporal acceleration compresses the available time for action realization, leading to incomplete or truncated behaviours. This highlights a deeper limitation: making time editable is necessary, but not sufficient, for long-horizon video generation.

Third, current evaluation metrics are limited in their ability to capture perceptual artifacts in intermediate motion, especially for articulated structures such as human limbs. In our experiments, we observe that models with improved visual consistency may receive worse scores on certain motion-based metrics due to limitations of the underlying detectors. This leads to a discrepancy between quantitative evaluation and perceived temporal quality.

Addressing these limitations will likely require combining explicit temporal representations with adaptive generation length and autoregressive or streaming extensions.
\section{Conclusion}

We introduced a temporal control framework for Diffusion Transformers based on explicit temporal factorization. Our key observation is that temporal controllability becomes limited when time is treated as a single overloaded variable. By separating time into global motion pacing and local temporal alignment, the proposed approach makes temporal dynamics directly editable rather than implicitly encoded in the diffusion process.

Explicit temporal conditioning improves intermediate motion consistency, reduces perceptual artifacts in articulated dynamics, and enables more coherent evolution of gradual physical processes. At the same time, our results show that improvements in temporal quality are not always reflected by standard evaluation metrics, highlighting a gap between quantitative measures and perceived motion realism.

Results on Wan2.2 and Kandinsky 5.0 Lite provide preliminary evidence that explicit temporal factorization may transfer across different DiT backbones despite differences in scale and output resolution. 

Overall, we argue that explicitly representing and factorizing time is an important step toward more physically consistent and reliable video generation.

\bibliography{refs}
\bibliographystyle{icml2026}

\newpage
\appendix
\onecolumn

\section{Related Work}\label{sec:related_works}

\paragraph{Adapter-based controllable diffusion.}
Recent work has shown that lightweight adapters are an effective way to extend pretrained diffusion backbones. Ctrl-Adapter~\cite{lin2025ctrladapter} and SimDA~\cite{xing2024simda} improve how external control signals are injected into image and video diffusion models, while FCA~\cite{liu2025fca} refines this direction through frame-wise text conditioning for video prediction. These methods make conditioning more flexible, but they do not directly address temporal controllability as a factorized problem. In particular, FCA adapts text conditioning at the frame level, whereas our method introduces an explicit decomposition of temporal control into a global rate signal and a local latent timeline.

\textbf{Motion control methods.}
Another line of work focuses on motion specification. Methods such as MotionPro~\cite{zhang2025motionpro}, MagicMotion~\cite{li2025magicmotion}, MotionAgent~\cite{liao2025motionagent}, and Trajectory Attention~\cite{xiao2024trajectoryattention} provide increasingly precise control over object paths, camera motion, and trajectory adherence. These approaches substantially improve controllable motion synthesis, but they still operate at the level of motion guidance: they determine \emph{what} moves and \emph{where} it moves, rather than making temporal progression itself editable.

\textbf{Temporal representation and decomposition.}
Work on temporal representation has also moved toward stronger factorization. VDT~\cite{lu2023vdt} established a transformer-based formulation for video diffusion with explicit spatial-temporal attention, while motion decomposition and factorized video generation methods~\cite{yu2024motiondecomposition,yu2024cmd} separate content and motion in the latent space to improve controllability and efficiency. TIC-FT~\cite{kim2025ticft} extends pretrained video diffusion models through temporally informed fine-tuning, but does not explicitly decompose time into global and local temporal variables. These works motivate our perspective that temporal control should be represented more explicitly, yet editable time is still not treated as the primary control target.

\textbf{Positioning of our method.}
Our method is complementary to these directions. Existing methods largely control motion; by contrast, we make time editable and treat control over dynamics as a natural consequence of temporal editability. Rather than replacing existing conditioning or motion-control modules, we move a pretrained DiT closer to temporal factorization while preserving the strengths of the original backbone.

\section{Detailed Method Formulation}
\label{app:formulation}

\textbf{Time-Conditioned Diffusion.}
As in Wan2.2~\cite{wan}, we adopt the same flow-matching formulation:
\begin{equation}
x_t = (1-t)z + t\epsilon, \qquad v^* = \epsilon - z,
\end{equation}
where $t \in [0,1]$ denotes the diffusion time and $\epsilon \sim \mathcal{N}(0,I)$.

We adopt the $v$-parameterization, where the target is defined as:
\begin{equation}
v^* = \epsilon - z.
\end{equation}

\textbf{Temporal Conditioning.}
We introduce a \emph{physical time} variable $\tau \in \mathbb{R}^{T}$ representing frame-wise temporal positions, and a global frame rate parameter $\mathrm{fps}$ controlling motion pacing.

Temporal conditioning is factorized into two components:
(i) a global time embedding
\[
E_{\mathrm{fps}}(\mathrm{fps}) = \mathrm{MLP}(\mathrm{Sinusoidal}(\mathrm{fps})),
\]
and
(ii) local temporal embeddings, where
\[
\tau_k = \left[0,\frac{1}{\mathrm{fps}},\dots,\frac{N_{\mathrm{latent}}}{\mathrm{fps}}\right]_k,
\]
\[
E_{\tau}(\tau_k) = \mathrm{MLP}(\mathrm{Sinusoidal}(\tau_k)).
\]

The model predicts:
\begin{equation}
v_{\theta}(x_t, \tau, \mathrm{fps}) = \mathrm{DiT}(\mathrm{PatchEmbed}(x_t) + E_{\tau}(\tau);\, e),
\end{equation}
where
\begin{equation}
e = E_t(t) + E_{\text{fps}}(\mathrm{fps})
\end{equation}
conditions the transformer via adaptive normalization, and positional encodings (e.g., RoPE~\cite{su2023roformerenhancedtransformerrotary}) capture spatial-temporal structure.

\textbf{Training Objective.}
The model is trained to match the velocity target:
\begin{equation}
\mathcal{L} = \| v_{\theta} - (\epsilon - z) \|_2^2.
\end{equation}

\section{Evaluation Metrics}
\label{app:metrics}

We report two benchmark metrics from VMBench, Perceptible Amplitude Score (PAS) and Motion Smoothness Score (MSS), together with three auxiliary temporal diagnostics: Flow Acceleration, LPIPS-Var, and Flow Directional Consistency.

\textbf{Perceptible Amplitude Score (PAS)}
We report PAS using the released VMBench evaluation implementation~\cite{vmbench2025}. The original VMBench paper defines PAS as a perceptual motion metric, whereas the released implementation computes PAS from foreground and background motion estimates obtained from CoTracker trajectories and segmentation masks.

Since PAS primarily measures motion amplitude rather than motion realism, it may not fully capture perceptual artifacts in articulated motion. We therefore interpret PAS as a motion-amplitude diagnostic and complement it with additional temporal-consistency metrics and human evaluation. Let $p_{t,j}$ be the tracked position of point $j$ at frame $t$, and let
\[
D=\sqrt{H^2+W^2}
\]
be the video diagonal. The normalized motion amplitude of a tracked region is
\[
\mathrm{MD}
=
\frac{1}{N}
\sum_{j=1}^{N}
\sum_{t=1}^{T-1}
\frac{\|p_{t+1,j}-p_{t,j}\|_2}{D}.
\]
In the released VMBench implementation, this motion degree is computed separately for the foreground subject and the background. The reported PAS value is then obtained as:
\[
\mathrm{PAS}=
\begin{cases}
\mathrm{MD}_{\mathrm{subj}}-\mathrm{MD}_{\mathrm{bg}}, & \text{if } \mathrm{MD}_{\mathrm{subj}}>\mathrm{MD}_{\mathrm{bg}},\\
\mathrm{MD}_{\mathrm{bg}}, & \text{otherwise.}
\end{cases}
\]
This formulation corresponds to the released VMBench evaluation code used in our experiments. Higher PAS indicates stronger perceptible motion.

\textbf{Motion Smoothness Score (MSS)}

MSS also follows VMBench~\cite{vmbench2025}. Let $s_t$ denote the Q-Align quality score of the sliding window centered at frame $t$. Temporal artifacts are identified by thresholding adjacent score differences:
\[
\mathcal{A}
=
\left\{
t \; \middle| \; |s_{t+1}-s_t| > \tau(\mathrm{PAS})
\right\},
\]
where $\tau(\mathrm{PAS})$ is an adaptive threshold determined by the motion amplitude regime. The final smoothness score is the proportion of non-artifact frames:
\[
\mathrm{MSS}
=
1-\frac{|\mathcal{A}|}{T}.
\]
Higher MSS indicates smoother temporal evolution and fewer perceptual breakdowns.

\textbf{Flow Acceleration}

To characterize short-term motion irregularity, we measure optical-flow acceleration. Let $F_t$ denote the mean optical-flow magnitude between frames $t$ and $t+1$. We define
\[
\Delta F_t = |F_{t+1}-F_t|,
\qquad
\mathrm{FlowAccel}
=
\frac{1}{T-2}\sum_{t=1}^{T-2}\Delta F_t.
\]
Lower values indicate more stable motion speed across time.

\textbf{LPIPS-Var}

To capture temporal instability at the perceptual level, we compute the standard deviation of adjacent-frame LPIPS:
\[
\mathrm{LPIPS\text{-}Var}
=
\sigma\!\left(\mathrm{LPIPS}(I_t, I_{t+1})\right),
\]
where $\sigma(\cdot)$ denotes the standard deviation over adjacent frame pairs. Lower values indicate more stable perceptual transitions and fewer short-term temporal disruptions.

\textbf{Flow Directional Consistency}

Flow Directional Consistency measures whether motion preserves a stable direction across adjacent time steps:
\[
\mathrm{FDC}
=
\frac{1}{T-2}
\sum_{t=1}^{T-2}
\cos(F_t, F_{t+1}),
\]
where the cosine similarity is computed between consecutive optical-flow vectors on moving regions. Higher values indicate stronger directional coherence.

\section{Benchmarks}
\label{app:benchmarks}

The Human Actions benchmark includes walking, running, dancing, waving, turning, and other articulated body motions. The Natural Processes benchmark includes rain, snow, waves, fog dissipation, cloud motion, and weather transitions. In total, we evaluate 80 generated videos at controlled target frame rates of 24, 30, and 60 FPS.
For each comparison setting (TA or FTTA against the baseline on Human Actions or Natural Processes), we evaluate 20 generated video pairs. Six raters provide pairwise judgments, resulting in 45 valid preference votes for each comparison setting and 180 votes in total across all Wan2.2 comparisons.

\subsection{Human Actions Benchmark}

We evaluate human-action generation using walking, running, jumping, dancing, body rotations, hand waving, and facial motion prompts. The main comparison is performed at 15, 24, 30, and 60 FPS. Since 15 FPS lies outside the training distribution and consistently produces weaker behaviour, the primary conclusions are drawn from 24--60 FPS.

\subsection{Natural Processes Benchmark}

We additionally evaluate rain, snow, waves, fog dissipation, cloud motion, wind-driven trees, coastline dynamics, and weather transitions. Unlike human actions, these scenarios require gradual continuous temporal evolution rather than discrete action timing, making long-range physical consistency the primary challenge.

\section{Side-by-Side Human Evaluation Details}

In addition to automatic metrics, we conduct a side-by-side human preference study on 80 generated videos. Six raters participate in the comparison, producing 180 votes in total. For each pair, raters choose one of four options: \textit{generation A}, \textit{generation B}, \textit{both good}, or \textit{both bad}. This protocol separates clear pairwise preference from cases where both results are acceptable or both fail perceptually.

We summarize pairwise preference using the win rate of generation A:
\begin{equation}
\mathrm{win\_rate}_A =
\frac{\mathrm{win}_A + \mathrm{win}_{\mathrm{both}}/2}
{\mathrm{win}_A + \mathrm{win}_B + \mathrm{win}_{\mathrm{both}}}.
\end{equation}
Here, $\mathrm{win}_A$ and $\mathrm{win}_B$ denote the number of votes in which A or B is preferred, and $\mathrm{win}_{\mathrm{both}}$ denotes the number of votes where both generations are judged good. Votes marked as \textit{both bad} are excluded from the denominator because they do not indicate a relative perceptual advantage for either model.

Each video pair is evaluated according to three criteria:

\textbf{Naturalness of dynamics} measures whether the observed motion looks natural and temporally plausible, without obvious visual or temporal artifacts.

\textbf{Consistency of dynamics with target speed} measures whether the generated motion matches the intended FPS regime, where 15 FPS corresponds to slow motion, 24--30 FPS to normal motion, and 60 FPS to fast motion.

\textbf{Prompt-action fidelity} measures whether the visible actions in the video match the actions explicitly described in the prompt.

For each comparison setting, we collect 45 pairwise votes from 6 raters over 20 generated video pairs. Each vote contains judgments for three criteria: naturalness, speed consistency, and prompt fidelity. We report win rates together with Wilson 95\% confidence intervals.
Table~\ref{tab:sbs_criteria_benchmarks_CI} reports Wilson 95\% confidence intervals for the SBS win rates.

\begin{equation}
\hat{p}_{\mathrm{Wilson}} =
\frac{
p + \frac{z^2}{2n}
}{
1 + \frac{z^2}{n}
}
\pm
\frac{
z
}{
1 + \frac{z^2}{n}
}
\sqrt{
\frac{p(1-p)}{n} + \frac{z^2}{4n^2}
},
\quad z=1.96.
\end{equation}

\begin{table}[t]
\centering
\caption{Side-by-side human evaluation against Wan2.2. Values denote win rates with Wilson 95\% confidence intervals. Values above 0.5 favor the time-controlled model.}
\label{tab:sbs_criteria_benchmarks_CI}
\small
\begin{tabular}{lcccc}
\toprule
& \multicolumn{2}{c}{Human Actions} & \multicolumn{2}{c}{Natural Processes} \\
\cmidrule(lr){2-3}\cmidrule(lr){4-5}
Criterion & TA & FTTA & TA & FTTA \\
\midrule
Naturalness & 0.58 [0.44, 0.71] & 0.45 [0.31, 0.59] & 0.43 [0.30, 0.57] & 0.50 [0.36, 0.64] \\
Speed consistency & 0.66 [0.51, 0.78] & 0.79 [0.65, 0.88] & 0.63 [0.48, 0.76] & 0.76 [0.62, 0.86] \\
Prompt fidelity & 0.45 [0.31, 0.59] & 0.37 [0.24, 0.52] & 0.43 [0.30, 0.57] & 0.69 [0.54, 0.81] \\
\bottomrule
\end{tabular}
\end{table}





\section{Training Setup}

The model is trained on a curated dataset of 40,000 videos covering multiple motion regimes. The data is distributed across three main FPS groups, 24--25 FPS (30\%), 30 FPS (30\%), and 60 FPS (40\%), with 60\% natural processes and physical events and 40\% human actions. This balance improves robustness across both articulated motion and continuous environmental dynamics.

We fine-tuned only the time-adapter parameters for 10,000 optimization steps on 1 GPU using AdamW (lr=1e-5, weight decay=1e-3, betas=(0.9,0.95), eps=1e-8), gradient clipping at 1.0, and EMA decay 0.999; the best model was selected at step 8,000. 

\end{document}